\documentclass{article}

\PassOptionsToPackage{numbers,compress}{natbib}
\usepackage[preprint]{neurips_2026}

\usepackage[utf8]{inputenc}
\usepackage[T1]{fontenc}
\usepackage{hyperref}
\usepackage{url}
\usepackage{booktabs}
\usepackage{amsfonts} 
\usepackage{amsmath}
\usepackage{nicefrac}
\usepackage{microtype}
\usepackage{xcolor}
\usepackage{graphicx}
\usepackage{enumitem}
\usepackage{float}

\title{Reasoning-aware Speculative Decoding for Efficient Vision-Language-Action Models in Autonomous Driving}

\author{%
  Anh-Dung Dinh\\
  School of Computer Science and Engineering \\
  UNSW Sydney \\
  \texttt{harry.dinh1@unsw.edu.au} \\
  \And
  Simon Khan \\
  Air Force Research Laboratory \\
  \texttt{simon.khan@us.af.mil} \\
  \And
  Flora D. Salim \thanks{Corresponding author.} \\
  School of Computer Science and Engineering \\
  UNSW Sydney \\
  \texttt{flora.salim@unsw.edu.au} \\
}

\begin{document}

\maketitle

\begin{abstract}

Modern Vision-Language-Action (VLA) planners for autonomous driving emit a chain-of-causation (CoC) reasoning step \emph{before} producing a trajectory. The reasoning is autoregressive and dominates inference latency, while the trajectory head is parallel and cheap. Latency is an operational constraint in autonomous driving, so accelerating the reasoning step is the central problem we address. We observe that CoC reasoning has two qualitatively different needs: most tokens continue routine setup that follows naturally from the ego-trajectory history, and a small fraction encode commitments that require fresh visual evidence about an unexpected situation. We split this reasoning into two specialized paths: a \emph{routine reasoner} that handles the predictable continuation by attending to trajectory history, and a \emph{deliberative reasoner} (the unmodified VLA target) that handles novel cases by attending to current visual evidence, using the speculative decoding framework as the architectural template for how the two paths cooperate. Unlike standard speculative decoding, our routine reasoner is not a smaller replica of the target; the two reasoners are deliberately specialized to read different parts of the prompt. We propose two techniques to realize this. First, we introduce \textbf{FlatRoPE}, a 1D rotary positional embedding in the draft that breaks the rotational symmetry of the target's 3D M-RoPE, redirecting attention away from visual tokens and onto trajectory-history tokens. Second, we introduce \textbf{Action-aware RL (AARL)}, a post-training stage that uses an action-quality reward together with a static-reference KL anchor. Together, our two-reasoner system reduces the reasoning-step running time by approximately $4\times$ relative to the original Alpamayo planner. 
\end{abstract}

\section{Introduction}
\label{sec:intro}

Modern Vision-Language-Action (VLA) planners for autonomous driving \cite{alpamayo,emma,drivelm} are sequential by construction: the model first emits a chain-of-causation (CoC) reasoning text (15--20 tokens) and \emph{then} produces a trajectory. The reasoning is autoregressive through an 8B+ multimodal backbone; the trajectory is produced in parallel by a flow-matching action head conditioned on the backbone's KV cache. The reasoning step is the bottleneck: it consumes an order of magnitude more wall-clock budget than the trajectory itself, and latency in driving translates directly into how soon a control commitment can be issued. Accelerating the reasoning step is therefore the central problem we address.

We observe that CoC reasoning has two different purposes. Most tokens continue routine setup. 
The next token follows naturally from the ego trajectory history (recent ego dynamics, partially-emitted CoC). A small fraction encode edge cases that require fresh visual evidence, such as a pedestrian half-stepping off a curb, an oncoming car crossing a center line -- where the next token cannot be predicted from history alone. In this work we split the reasoning step into two paths specialized for these two needs: a \emph{routine reasoner} that handles predictable continuation by attending to trajectory history, and a \emph{deliberative reasoner} that handles the novel cases by attending to current visual evidence. The routine reasoner is fast and the deliberative reasoner is heavier; we use the speculative decoding framework \cite{leviathan,chen2023} as the architectural \emph{template} for how they cooperate -- the routine reasoner proposes the next $K$ tokens, the deliberative reasoner verifies them in one pass, and the longest matching prefix is committed.

Critically, our routine reasoner is \emph{not} a reduced-parameter distillation of the target model. The two reasoners are deliberately specialized to read \emph{different} parts of the prompt: the routine reasoner attends primarily to trajectory-history positions where the next routine token can be predicted, while the deliberative reasoner is the unmodified VLA target and retains its full visual attention to detect novelty. The role of each is set by what kind of token-prediction it is good at, not by parameter count. The mechanism that produces this specialization on the draft side is a position-encoding change we call \textbf{FlatRoPE}: the natural choice for a multimodal draft -- mirroring the target's 3D M-RoPE. This couples the draft's attention to the visual token cluster and diverts the attention budget away from the trajectory-history positions where routine signal lives (Section~\ref{sec:method-1d}). FlatRoPE replaces 3D M-RoPE with plain 1D rotary, breaking the rotational symmetry of adjacent vision patches and redirecting the attention budget onto trajectory history. We further introduce a reinforcement-learning post-training stage (Section~\ref{sec:method-rl}) with a three-component reward and a static-reference KL anchor that fixes a self-removal pathology in the standard policy anchor. Our contributions:

\begin{enumerate}[noitemsep]
\item \textbf{FlatRoPE}: plain-1D rotary in the draft, motivated by an attention-entropy diagnostic showing that 3D M-RoPE clusters vision tokens and disproportionately reduces attention mass on trajectory-history tokens. FlatRoPE is the mechanism that produces the routine-reasoner-vs-deliberative-reasoner specialization on the draft side. Across two draft architectures and two open VLA targets, FlatRoPE consistently outperforms the 3D-mirroring baseline on accepted length.
\item \textbf{AARL post-training} with a static-reference KL anchor. The standard policy anchor self-removes as the draft drifts; the static reference does not. The result is the first post-training method we are aware of that improves both validation \emph{and} held-out test accepted length on an open AV VLA.
\item \textbf{Empirical validation that this specialization fits autonomous driving.} We measure per-token attention budget at the draft's input fusion (showing FlatRoPE redirects attention to trajectory history as designed), provide a token-level routing diagnostic on real CoC traces (showing the routine reasoner handles setup tokens autonomously while the deliberative reasoner is selectively invoked at visually-grounded decision points), and quantify end-to-end latency reduction on val, on-shelf, and off-shelf test splits.
\end{enumerate}

In summary, these three contributions produce a two-mode decoder whose routine path handles the majority of CoC tokens without invoking the deliberative path. With $L=4.6$ at block size 16 the routine path handles $\sim 78\%$ of CoC tokens, giving a theoretical wall-clock speedup of $3.5\times$ at iso-quality on Alpamayo-R1.

\section{Related work}
\label{sec:related}

\textbf{Vision-language-action models for driving.}
End-to-end VLA models for autonomous driving have rapidly converged on a common
recipe: a vision-language backbone consumes multi-camera images and ego state,
and an action head produces a future trajectory. EMMA \cite{emma} demonstrated
end-to-end driving from a Gemini-class backbone; DriveLM \cite{drivelm}
introduced graph-structured visual question answering as a planning interface.
DriveGPT4 \cite{drivegpt4} showed that a multimodal LLM can be fine-tuned to
emit interpretable driving rationales. 

\begin{figure}[htpb!]
    \centering
    \includegraphics[width=1.0\linewidth]{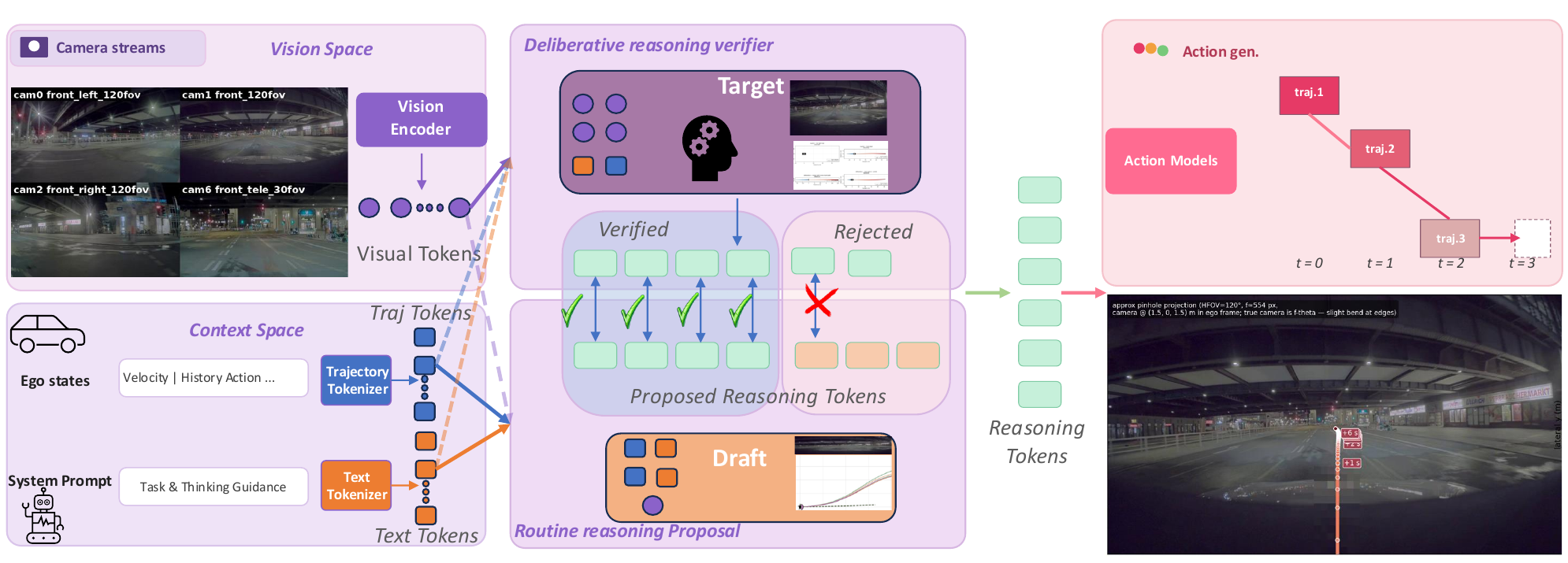}
    \caption{\textbf{Reasoning-Aware Speculative Decoding Framework}: The target model encodes visual features and trajectory data into tokens, which a compact draft model then uses to generate reasoning proposals. While the target model maintains a broad visual focus, the draft model prioritizes historical trajectory patterns to perform lightweight, routine-based reasoning. In the event of a novel scenario, any disagreement between the models triggers a rejection by the target, initiating a new round of proposals to ensure accuracy.}
    \label{fig:hook}
\end{figure}

World-model-grounded VLAs such as
GAIA-1 \cite{gaia1} and Wayve's LINGO/UniAD-style stacks
\cite{lingo,uniad} couple language reasoning with imagined futures.
RoboTron-Drive \cite{robotron} and OmniDrive \cite{omnidrive} focus on
perception-grounded reasoning, while OpenEMMA \cite{openemma} and
MiniDrive \cite{minidrive} aim at compact, deployable variants.
Alpamayo-R1 \cite{alpamayo} and AutoVLA \cite{autovla} -- the two open
targets we evaluate against -- additionally emit a chain-of-causation (CoC)
reasoning step before invoking a continuous flow-matching or diffusion
trajectory head, making the autoregressive language portion the dominant
inference cost. Our work targets exactly that bottleneck.

\textbf{Efficiency for autonomous-driving reasoning.}
Two complementary directions have emerged. The first is \emph{architectural
compaction}: distilling the full VLA into a smaller backbone
(MiniDrive \cite{minidrive}, TinyDrive \cite{tinydrive}), pruning attention
heads, or quantizing weights and activations for on-vehicle deployment.
The second is \emph{reasoning-aware inference}: limiting the chain-of-thought
to high-uncertainty scenarios via early exit \cite{deer-vlm} or selective
deliberation \cite{drivegpt4-think}, caching prompt prefixes across frames
\cite{driving-cache}, or skipping the language step entirely on routine clips
\cite{action-only-mode}. Both reduce \emph{when} the deliberative model runs
but do not change its per-token cost. Speculative decoding is orthogonal:
it preserves the target's exact greedy outputs while spreading the per-step
cost across cheap routine-reasoner forwards. To our knowledge, no prior work
has applied speculative decoding to AV CoC reasoning.

\textbf{Speculative decoding and parallel-decoding methods.}
\citet{leviathan} and \citet{chen2023} introduced speculative decoding for
autoregressive language models. The two main subsequent design axes have been
(i) the draft architecture and (ii) the verification topology. On the draft
side, Medusa \cite{medusa} adds parallel prediction heads to the target
itself; EAGLE/EAGLE-2/EAGLE-3 \cite{eagle,eagle2,eagle3} train an
autoregressive draft conditioned on multi-layer target hidden states;
SpS \cite{sps}, BiLD \cite{bild}, and KV-Reuse drafts \cite{kvreuse} explore
encoder-side acceleration; block-diffusion drafts \cite{block-diffusion,sahoo2024}
predict groups of tokens in parallel via masked-denoising objectives. On the
verification side, tree-decoding variants \cite{specinfer,medusa,eagle2} verify
multiple candidate paths in a single target forward via custom attention masks,
trading more verify-time compute for higher accept rate. Lookahead decoding
\cite{lookahead} eliminates the draft entirely by re-using the target itself
as a noisy denoiser. Adjacent work compresses or shares the target's KV cache
across speculative iterations \cite{kvshare,gqa}. All of these target
text-only chat or code completion. None have studied the position-encoding
choice for drafts of \emph{multimodal} targets, where the target's native
M-RoPE 3D structure couples the draft's attention to the visual cluster -- the
phenomenon we identify and fix in this paper.

\textbf{Two-mode reasoning in foundation models.}
A recurring design pattern in foundation-model inference is to route easy
queries past the deliberative-reasoning path while preserving its outputs:
chain-of-thought decomposition with selective rollouts \cite{system15} and
self-consistency \cite{self-consistency} both follow this pattern at the
query level. Standard speculative decoding applies the same routing at the
token level, but with a particular instantiation: the draft is a small
imitator of the target, trained to match the target's distribution. We
depart from that recipe. Our routine reasoner is not a small target; it is
specialized to attend to trajectory history, while the deliberative
reasoner (the unmodified VLA target) retains its full visual attention.
The two reasoners are good at different things by construction, and the
speculative-decoding loop is the protocol that lets them cooperate. We
make this specialization explicit in the AV setting and motivate where
the routine reasoner's attention budget should land -- on the small
trajectory-history slice where routine signal lives, not on the visual
blob that the routine tokens never need to inspect.

\section{Background}
\label{sec:background}

We define the
two-stage VLA driving target, speculative decoding, and the multimodal
rotary position encoding used by the targets we evaluate. 
Our specific design choices is deferred to Section~\ref{sec:method}.

\subsection{Vision-language-action driving with reasoning}
\label{sec:bg-vla}

\textbf{Inputs.} The target VLA, parameterized by $\theta_T$, consumes a
multi-camera observation $\mathbf{I} \in \mathbb{R}^{C \times H \times W}$
(stacked $C$ views), an ego-state / history vector
$\mathbf{e} \in \mathbb{R}^{d_e}$, and a textual prompt
$\mathbf{P} \in \mathcal{V}^{|\mathbf{P}|}$ over the target tokenizer's
vocabulary $\mathcal{V}$. We collect them as $x = (\mathbf{I}, \mathbf{e}, \mathbf{P})$.

\textbf{Two-stage output.} The target produces a rationale
$\mathbf{r} \in \mathcal{V}^{n}$ -- a chain-of-causation (CoC) for
Alpamayo-R1, a chain-of-thought (CoT) for AutoVLA -- followed by an
action trajectory
$\boldsymbol{\tau} \in \mathbb{R}^{T_f \times d_a}$ over a horizon of
$T_f$ control steps in $d_a$-dimensional action space (typically
$d_a = 2$ for longitudinal acceleration and lateral curvature).

\textbf{Stage 1 (autoregressive reasoning).} The rationale is generated
token by token by an autoregressive language model:
\begin{equation}
\mathbf{r}_t \sim \pi_T(\cdot \mid x, \mathbf{r}_{1:t-1}; \theta_T),
\qquad t = 1, \dots, n,
\label{eq:stage1}
\end{equation}
terminating at a model-specific stop token.

\textbf{Stage 2 (action generation).} Conditioned on $x$ and the committed
rationale $\mathbf{r}$, the target emits the trajectory through a
model-specific function $g_T$:
\begin{equation}
\boldsymbol{\tau} = g_T\!\left(x, \mathbf{r}; \theta_T\right).
\label{eq:stage2}
\end{equation}
The two targets instantiate $g_T$ differently. \textbf{Alpamayo-R1}
produces $\boldsymbol{\tau}$ via a flow-matching diffusion head reading the
backbone's KV cache after position $|\mathbf{P}| + n$:
\begin{equation}
g_T^{\text{Alpamayo}}\!\left(x, \mathbf{r}\right) =
\mathrm{ODE}\!\left[
v_{\theta_T}\!\big(\boldsymbol{\tau}_s, s; \, \mathrm{KV}(x, \mathbf{r})\big)
\right]_{s=0 \to 1},
\label{eq:alpamayo-stage2}
\end{equation}
where $v_{\theta_T}$ is a learned vector field, $s$ is the diffusion time,
and $\boldsymbol{\tau}_0 \sim \mathcal{N}(0, I)$. AutoVLA emits the
trajectory as a discrete token sequence decoded autoregressively from the
same backbone:
\begin{equation}
g_T^{\text{AutoVLA}}\!\left(x, \mathbf{r}\right) =
\mathrm{detokenize}\!\left(\mathbf{a}_{1:T_f}\right),
\quad
\mathbf{a}_t \sim \pi_T\!\left(\cdot \mid x, \mathbf{r}, \mathbf{a}_{1:t-1}; \theta_T\right).
\label{eq:autovla-stage2}
\end{equation}

\textbf{Latency budget.} Let $C_T$ be the per-token cost of one target
forward pass and $C_g$ the cost of $g_T$. The total inference cost is
\begin{equation}
\underbrace{n \cdot C_T}_{\text{Stage 1}} \,+\,
\underbrace{C_g}_{\text{Stage 2}}.
\label{eq:cost}
\end{equation}

\subsection{Speculative decoding}
\label{sec:bg-spec}

A draft $D$ with parameters $\theta_D$ defines a surrogate token
distribution $\pi_D(\cdot \mid x, \mathbf{r}_{1:t-1}; \theta_D)$ over
$\mathcal{V}$. Speculative decoding replaces Equation~\ref{eq:stage1} with
an iterative \emph{propose / verify / commit} loop. Let $i$ index the loop
iteration, $\mathbf{r}_{1:t_i}$ denote the prefix already committed, and
$K$ a block size.

\textbf{Propose.} The draft emits $K$ candidate tokens in a single forward:
\begin{equation}
\hat{\mathbf{r}}_{t_i+1:t_i+K} =
\mathrm{Propose}_K\!\left(\pi_D, x, \mathbf{r}_{1:t_i}\right).
\label{eq:propose}
\end{equation}

\textbf{Verify.} The target evaluates them in one batched forward pass:
\begin{equation}
\mathbf{r}^{\star}_{t_i+j} =
\arg\max_{v \in \mathcal{V}} \pi_T\!\big(v \mid x,\, \mathbf{r}_{1:t_i},\,
\hat{\mathbf{r}}_{t_i+1:t_i+j-1};\, \theta_T\big),
\quad j = 1, \dots, K.
\label{eq:verify}
\end{equation}

\textbf{Commit.} The longest matching prefix is the accepted length:
\begin{equation}
\ell_i \;=\; \max\Big\{ l \in \{0, \dots, K\} :
\hat{\mathbf{r}}_{t_i+j} = \mathbf{r}^{\star}_{t_i+j} \;\forall j \le l \Big\},
\label{eq:accept}
\end{equation}
and the iteration commits $\ell_i + 1$ tokens (the matched prefix plus a
``bonus'' from the target at position $\ell_i + 1$):
\begin{equation}
\mathbf{r}_{t_i+1:t_i+\ell_i+1} \,=\,
(\hat{\mathbf{r}}_{t_i+1}, \dots, \hat{\mathbf{r}}_{t_i+\ell_i},
 \mathbf{r}^{\star}_{t_i+\ell_i+1}),
\qquad t_{i+1} \,=\, t_i + \ell_i + 1.
\label{eq:commit}
\end{equation}
The loop terminates when a stop token is committed.

\textbf{Mean accepted length and theoretical speedup.}
\begin{equation}
L \;=\; \mathbb{E}_{i}\!\left[ \ell_i + 1 \right] \;\in\; [1, K+1],
\qquad
\mathcal{S} \;\approx\; \frac{L \cdot C_T}{C_T^{(K)} + C_D^{(K)}},
\label{eq:L-and-speedup}
\end{equation}
where $C_T^{(K)}$ and $C_D^{(K)}$ are the per-iteration target-verify and
draft-propose costs.

\subsection{Multimodal rotary position encoding (M-RoPE)}
\label{sec:bg-mrope}

The targets we evaluate use a multimodal extension of rotary position
encoding (M-RoPE) \cite{qwenvl} on every attention layer.

\paragraph{1D rotary.} Let $\mathbf{q} \in \mathbb{R}^{h}$ be a per-head
query at sequence position $m \in \mathbb{N}$ with $h$ even, and let
$\boldsymbol{\omega} \in \mathbb{R}^{h/2}$ be a fixed frequency vector
(typically $\omega_d = b^{-2d/h}$ for a base $b$, e.g.\ $b = 10{,}000$).
Standard 1D rotary embedding pairs adjacent dimensions and rotates each
pair by an angle proportional to position:
\begin{equation}
\mathrm{RoPE}(\mathbf{q}, m) \;=\; \mathbf{R}(m, \boldsymbol{\omega})\, \mathbf{q},
\qquad
\mathbf{R}(m, \boldsymbol{\omega}) =
\bigoplus_{d=0}^{h/2 - 1}
\begin{pmatrix} \cos(m \omega_d) & -\sin(m \omega_d) \\
                \sin(m \omega_d) & \cos(m \omega_d) \end{pmatrix},
\label{eq:rope-1d}
\end{equation}
where $\bigoplus$ denotes block-diagonal concatenation.

\paragraph{M-RoPE 3D.} For multimodal sequences mixing text and image
patches, the target assigns each token a three-axis position
$\mathbf{m} = (m^T, m^H, m^W) \in \mathbb{N}^3$ -- temporal, height, and
width coordinates within the camera-frame grid. The head dimensions are
partitioned by an \texttt{mrope\_section} vector $(s^T, s^H, s^W)$ with
$s^T + s^H + s^W = h/2$. M-RoPE applies a separate rotation on each
section driven by the corresponding axis position:
\begin{equation}
\mathrm{MRoPE}(\mathbf{q}, \mathbf{m}) \;=\;
\begin{bmatrix}
\mathbf{R}_{s^T}(m^T,\, \boldsymbol{\omega}^T)\, \mathbf{q}^{(T)} \\
\mathbf{R}_{s^H}(m^H,\, \boldsymbol{\omega}^H)\, \mathbf{q}^{(H)} \\
\mathbf{R}_{s^W}(m^W,\, \boldsymbol{\omega}^W)\, \mathbf{q}^{(W)}
\end{bmatrix},
\label{eq:rope-3d}
\end{equation}
where
$\mathbf{q} = [\mathbf{q}^{(T)};\, \mathbf{q}^{(H)};\, \mathbf{q}^{(W)}]$
is the head split along the feature dim, $\mathbf{R}_{s}$ is the
$s$-pair block-rotation of Equation~\ref{eq:rope-1d} restricted to $2s$
dimensions, and
$(\boldsymbol{\omega}^T, \boldsymbol{\omega}^H, \boldsymbol{\omega}^W)$
are the frequency vectors of the three axes; in practice each axis is
allocated only a fraction of the head dimension (concrete values for the
backbone we use are given in Section~\ref{sec:experiments}). For text
tokens the three coordinates
collapse to the sequence index, $m^T = m^H = m^W = m$, and
Equation~\ref{eq:rope-3d} reduces to Equation~\ref{eq:rope-1d}. For image
patches the three coordinates encode the frame and grid coordinates of
the patch.

\section{Method}
\label{sec:method}

\subsection{The two-mode decoder: specialized draft and visual target}

Building on Section~\ref{sec:bg-spec}, we instantiate the routine reasoner
$D$ as a small transformer that ingests the target's hidden states at one
or more layers as conditioning, and produces
$\pi_D(\cdot \mid x, \mathbf{r}_{1:t})$. The hidden-state conditioning is
deliberate, not incidental: the routine reasoner has only a small
parameter budget (well under 1B), far too small to reproduce the target's
visual encoder from scratch. Re-running visual reasoning end-to-end inside
$D$ would either require scaling $D$ until it is no longer ``small'' (and
spec decoding's speedup vanishes), or accepting a degraded visual
representation. Instead we let $D$ \emph{reuse} the visual understanding
the target has already computed: target hidden states, sliced from one or
more layers, are fed into $D$ as cross-attention K/V context. The target
does the heavy visual work once; $D$'s parameters are spent on combining
that context with the trajectory-history positions to predict the next
routine token. Unlike a generic spec-decode draft trained to mimic the
target's distribution at every position, $D$ is specialized: its attention
is steered toward trajectory-history positions (recent ego dynamics + the
partially-emitted CoC) rather than the visual cluster. The mechanism that
produces this specialization is the position-encoding choice in
Section~\ref{sec:method-1d}; the consequence is that $D$ is fast at
predicting routine tokens (where trajectory history is sufficient) and
intentionally weak on novel-situation tokens (where fresh visual evidence
is required). The deliberative reasoner is the unmodified VLA target with
its native 3D M-RoPE intact; it retains its full visual attention for
verification and for emitting the bonus token when the routine reasoner
is rejected. Two draft architecture families have established themselves
in the spec-decoding literature and we evaluate both: a \emph{block
diffusion} draft that computes $\hat{\mathbf{r}}_{t+1:t+K}$ via a single
masked-denoising step, and an \emph{autoregressive} draft with multi-layer
feature fusion that emits $K$ tokens via a depth-bounded chain. The
specific architectures we instantiate (sizes, layer counts, fusion
schemes) and the references they correspond to in the literature are
deferred to Section~\ref{sec:experiments}. The trajectory head $g_T$ is
unchanged from Equations~\ref{eq:alpamayo-stage2}-\ref{eq:autovla-stage2};
only Stage~1 of Equation~\ref{eq:cost} is replaced by the two-mode loop.

\subsection{FlatRoPE: from 3D M-RoPE to 1D in the draft}
\label{sec:method-1d}

\paragraph{Background: target's 3D M-RoPE.}
The multimodal language-model backbones used by current VLA targets
apply 3D M-RoPE with hyperparameters
$\texttt{mrope\_section} = (s_T, s_H, s_W)$ summing to the rotary half-dim
$D_r = D_{\text{head}}/2$, and a boolean \texttt{interleaved}. Concretely,
for each rotary-pair index $d \in \{0,\dots,D_r-1\}$ we define an axis
assignment $a(d) \in \{T,H,W\}$:
\begin{equation}
a(d) =
\begin{cases}
T, & 0 \le d < s_T,\\
H, & s_T \le d < s_T + s_H,\\
W, & s_T + s_H \le d < D_r,
\end{cases}
\quad\text{(blocked)},
\qquad
a(d) = \{T,H,W\}_{(d \bmod 3)}
\quad\text{(interleaved)}.
\label{eq:axis-assign}
\end{equation}
The specific MLLM family we evaluate against (along with concrete values
of $\texttt{mrope\_section}$ and the \texttt{interleaved} flag) is
deferred to Section~\ref{sec:experiments}; the analysis below holds for
any 3D-M-RoPE backbone with $s_T, s_H, s_W$ each substantially smaller
than $D_r$. Each token $i$ carries a 3D coordinate
$\big(p_T(i), p_H(i), p_W(i)\big)$, and the rotation angle at rotary-pair
$d$ is
\begin{equation}
\phi_d(i) = \theta_d \, p_{a(d)}(i),
\qquad
\theta_d = 10000^{-2d/D_{\text{head}}}.
\label{eq:mrope-phase}
\end{equation}
For text and ego-history tokens, all three coordinates collapse to the
sequence index $p_T = p_H = p_W = m$, so $\phi_d(i) = \theta_d \, m(i)$ at
every $d$ and 3D M-RoPE reduces to 1D RoPE. For vision patches the three
coordinates are independent grid offsets, and the rotation is genuinely
three-dimensional.

\textbf{Why mirroring is harmful.}
A natural draft design ingests target hidden states (already rotated by
target M-RoPE) and applies the same 3D M-RoPE on its own queries -- mirroring
$(s_T, s_H, s_W)$ and \texttt{interleaved}. Two adjacent vision patches at
$(p_T, p_H, p_W)$ and $(p_T, p_H, p_W{+}1)$ differ only in the
$s_W$-dim subspace of $\phi$ (similarly for height-adjacent patches in the
$s_H$ subspace). Because each spatial axis is allocated only a fraction
$s_W / D_r$ (or $s_H / D_r$) of the rotary dimensions -- typically less
than one third in current MLLM backbones -- the post-rotation Q/K differ
over only that small subspace and remain near-aligned in the rest. The
dot-product after rotation collapses onto the cluster, the softmax cannot
resolve individual patches, and the draft's attention budget pools on the
visual cluster as a whole.

\paragraph{FlatRoPE projection.}
We replace the draft's 3D rotation with a 1D rotation indexed by the
sequence position only. For every token $i$ in the draft input -- vision,
text, ego-history alike -- we set
\begin{equation}
p_T^{\text{flat}}(i) = p_H^{\text{flat}}(i) = p_W^{\text{flat}}(i)
\;=\; m(i) \in \{0,1,\dots,T_{\text{seq}}-1\}.
\label{eq:flatrope-proj}
\end{equation}
Substituted into Equation~\ref{eq:mrope-phase}, the per-dim phase becomes
$\phi_d^{\text{flat}}(i) = \theta_d \, m(i)$, identical for every $d$
\emph{regardless of the axis assignment $a(d)$}. Two important consequences
follow.

\textit{(i)} \textbf{$\texttt{mrope\_section}$ and $\texttt{interleaved}$ are
structurally irrelevant under FlatRoPE.} Both \emph{blocked} and
\emph{interleaved} axis assignments in Equation~\ref{eq:axis-assign}
collapse to the same scalar phase $\theta_d m(i)$, since the axis-selected
coordinate is identical across $T,H,W$. Mismatch between the draft's
declared $\texttt{mrope\_section}$ and the target's no longer affects the
draft's queries or keys. We retain only the target's $\theta_d$ schedule
(same head dim, same base $10000$) so target-derived hidden states and
draft-side queries share frequency tables; the position projection itself
is $\mathbb{R}^3 \to \mathbb{R}$ via $(p_T, p_H, p_W) \mapsto m$.

\textit{(ii)} \textbf{The asymmetry between target and draft is intentional.}
The target's hidden states are still produced under the spatially-aware
3D rotation (the target is unmodified); but the draft attends to them
with queries rotated by 1D phases. After FlatRoPE projection, two
vision patches with sequence positions $m_i \ne m_j$ differ in rotation
across \emph{every} rotary-pair $d$ by $\theta_d(m_j - m_i)$, not just
the $s_W$ or $s_H$ subspace. Q/K post-rotation are well separated, the
softmax discriminates between patches, and the freed budget redistributes
onto trajectory-history positions where the next routine token lives.

\paragraph{Empirical confirmation.}
We measure the consequence on the layer-0 cross-attention pattern of a
converged warm-SFT block-diffusion draft (Table~\ref{tab:attn-budget}). Visual
tokens occupy $\sim$96\% of the prompt sequence; trajectory and committed-CoC
tokens are the remaining 4\%. Under 3D M-RoPE the softmax pools mass on
the visual cluster (84.5\% of total mass) and starves trajectory tokens
(7.1\%) and the draft's own generated tokens (8.4\%). FlatRoPE pulls the
visual mass down to 70.8\%, restoring the freed budget to trajectory
tokens (8.4\%) and generated tokens (20.8\%) -- exactly the positions where the
next routine-token signal lives. Within-trajectory-and-text attention
entropy correspondingly rises from 0.404 to 0.540, indicating the softmax
is now discriminating among trajectory and text positions rather than
concentrating on a single sink token.

\begin{table}[t]
\centering
\caption{Attention budget at layer-0 cross-attention of a converged block-diffusion draft, averaged over 32 heads and one test clip. 3D M-RoPE drains attention from the action-relevant trajectory and generated tokens onto the visual cluster; 1D restores it.}
\label{tab:attn-budget}
\begin{tabular}{lcc}
\toprule
Token type & 1D rotary (ours) & 3D M-RoPE (baseline) \\
\midrule
visual tokens                          & 0.708 & \textbf{0.845} \\
trajectory tokens                      & \textbf{0.084} & 0.071 \\
generated tokens                       & \textbf{0.208} & 0.084 \\
\midrule
entropy within trajectory and text (norm.) & \textbf{0.540} & 0.404 \\
\bottomrule
\end{tabular}
\end{table}

\subsection{Action-aware RL post-training}
\label{sec:method-rl}

\paragraph{Problem formulation.}
After supervised fine-tuning, we want to further train the draft to
maximize the realized speedup. From Equation~\ref{eq:L-and-speedup}, this is
bounded by the mean accepted length. Define a per-prompt proposal
$\hat{\mathbf{r}}_{1:K} \sim \pi^\theta_D(\cdot \mid x)$ and its
acceptance length against the target's greedy continuation $\mathbf{r}^\star$:
\begin{equation}
\ell\big(\hat{\mathbf{r}}_{1:K};\, x\big)
= \max\big\{\, k \in \{0,\dots,K\} \,:\,
\hat r_j = r^\star_j ~\forall j \le k \,\big\}.
\label{eq:accept-length}
\end{equation}
The post-training objective is to find draft parameters $\theta$ that
maximize the expected acceptance length:
\begin{equation}
\theta^\star \;=\; \arg\max_{\theta}\;
\mathbb{E}_{x \sim \mathcal{D}}\,
\mathbb{E}_{\hat{\mathbf{r}} \sim \pi^\theta_D(\cdot \mid x)}
\big[\, \ell(\hat{\mathbf{r}};\, x) \,\big].
\label{eq:aarl-objective}
\end{equation}

\paragraph{Why RL.}
Equation~\ref{eq:aarl-objective} is non-differentiable in $\theta$: the
acceptance length is a $\max$ over hard token-level argmax matches, with
no smooth surrogate. The standard supervised proxy -- per-position
cross-entropy of $\pi^\theta_D$ against $\mathbf{r}^\star$ -- treats every
position uniformly and ignores the structural fact that an error at
position $k$ erases all gains at positions $> k$. Two well-known
consequences are: (i) two policies with the same average per-position CE
can have quite different $L$, depending on how their errors distribute
across positions; and (ii) once SFT has converged, lowering CE further
yields almost no $L$ improvement (the residual CE comes mostly from
genuinely uncertain positions). We therefore treat
Equation~\ref{eq:aarl-objective} directly as a single-step contextual
bandit, sample $K$ rollouts per prompt, and apply group-relative policy
optimization (GRPO) to estimate the gradient of $\mathbb{E}[\ell]$.

\paragraph{Action-aware reward.}
The acceptance length $\ell$ itself is a sparse 0/1-style signal -- it
provides no gradient until at least one position matches and it does not
distinguish among rollouts that fail at the same position. We therefore
replace $\ell$ with a denser proxy reward grounded in the trajectory the
target ultimately produces. For each prompt $x$ we sample $K=5$ rollouts
(stochastic only at greedy-rejected positions; greedy-matched positions
are frozen at the deterministic argmax to reduce variance). For the
$k$-th rollout we form a contaminated rationale $\tilde{\mathbf{r}}_k$
by replacing the first $N$ rejected positions of $\mathbf{r}^\star$ with
the $k$-th sample, run the target's flow-matching head $g_T$ to obtain
$\boldsymbol{\tau}^{\text{mix}}_k = g_T(x, \tilde{\mathbf{r}}_k)$, and
compute
\begin{equation}
R_k \;=\; w_\text{traj}\,R^{\text{traj}}_k \;+\; w_\text{text}\,R^{\text{text}}_k,
\qquad
R^{\text{traj}}_k = -\big\|\boldsymbol{\tau}^{\text{mix}}_k - \boldsymbol{\tau}^\star\big\|_2^2,
\quad
R^{\text{text}}_k = \frac{1}{N}\sum_{j \in \mathcal{R}_k^{[:N]}}
\mathbb{1}\!\big[\hat r_{k,j} = r^\star_j\big],
\label{eq:reward}
\end{equation}
where $\boldsymbol{\tau}^\star = g_T(x, \mathbf{r}^\star)$ is the
target's greedy trajectory and $\mathcal{R}_k^{[:N]}$ are the substituted
positions. The reward is `action-aware' in the sense that token-level
disagreements with the target are penalized in proportion to how much
they perturb the target's eventual trajectory, giving the draft direct
gradient signal toward the positions whose token choice actually matters
for downstream control.

\paragraph{GRPO update.}
The advantage is the group-relative baseline
$\hat A_k = R_k - \bar R$ with $\bar R = \tfrac{1}{K}\sum_k R_k$. The
policy gradient accumulates only over the sampled (rejected) positions:
greedy positions have zero log-prob gradient by construction. Combined
with the static-reference KL anchor (below), the per-prompt loss is
\begin{equation}
\mathcal{L}(\theta) \;=\;
- \frac{1}{K}\sum_{k=1}^{K}
\hat A_k \!\!\sum_{j \in \mathcal{R}_k}
\log \pi_\theta\!\big(\hat r_{k,j} \,\big|\, x, \tilde{\mathbf{r}}_{k,<j}\big)
\;+\;
\beta \cdot \frac{1}{|\mathcal{A}_{\text{ref}}|}
\sum_{j \in \mathcal{A}_{\text{ref}}}
\mathrm{KL}\!\big(\pi_\theta(\cdot \mid x, \cdot)\,\|\,\pi_{\text{ref}}(\cdot \mid x, \cdot)\big),
\label{eq:aarl-loss}
\end{equation}
with $\beta=0.02$ in our runs.

\paragraph{Static-reference anchor.}
The KL term in Equation~\ref{eq:aarl-loss} is gated by an anchor set
$\mathcal{A} \subseteq \{1,\dots,K\}$. The standard recipe defines the
anchor against the \emph{live} policy:
$\mathcal{A}_{\text{policy}}(\theta) = \{j : \arg\max \pi_\theta(\cdot|...) = r^\star_j\}$.
This gating self-removes during drift: as $\theta$ updates, a position
whose argmax flips from correct to wrong is reclassified out of the
anchor set, exactly when the regularizer would have corrected the
flip. Empirically this manifests as $\texttt{draft\_final}$ regressing
relative to mid-training checkpoints. We instead use a \emph{static}
reference set
$\mathcal{A}_{\text{ref}} = \{j : \arg\max \pi_{\text{ref}}(\cdot|...) = r^\star_j\}$
fixed at the start of post-training -- positions where the frozen
SFT-init was already correct receive permanent anchoring, positions
where it was wrong remain free to move under the policy gradient. The single-line code-level diff can be found in appendix.
The reference logits are produced by a frozen copy of the SFT draft in
the same forward pass, contributing negligible additional cost. We find
no benefit from periodically refreshing the reference, consistent with
the analysis above: the pathology is the live-policy coupling itself, not
the staleness.

\section{Experiments}
\label{sec:experiments}


\textbf{Targets and drafts.} We evaluate two open AV VLA targets. \textbf{Alpamayo-R1} \cite{alpamayo} is 10B parameters with an 8B Qwen3-VL backbone and a flow-matching trajectory head; the autoregressive component emits CoC reasoning. \textbf{AutoVLA} \cite{autovla} is an interesting work, but by the time we make submission, AutoVLA has not released the checkpoint yet. As a result, we will keep the extension of our work to AutoVLA in the future works. We pair each target with two drafts: \textbf{DFlash} \cite{block-diffusion}, a small block-diffusion network operating on multi-layer target hidden states with block size $K=16$; and \textbf{EAGLE-3} \cite{eagle}, an autoregressive 1-layer draft with multi-layer feature fusion (target hiddens at layers $\{1, 17, 32\}$ for the 36-layer Qwen3-VL backbone) and a 7-step training rollout per the EAGLE-3 paper recipe. For each (target, draft) pair we train two variants -- 3D M-RoPE rotary in the draft (the natural baseline) and 1D arange rotary (ours) -- with the same loss and hyperparameters.

\textbf{Data.} Alpamayo-R1: 11,655 training clips with cached target greedy CoC outputs (mean output length 17.4 tokens). Held-out splits: val (300), on-shelf test (200), off-shelf test (290 from a separate annotation source). 

\textbf{Metric.} Mean accepted length $L$ at $K{=}16$ verification, evaluated greedily at temperature $0$. Higher is better.

\subsection{Main result: FlatRoPE across architectures and targets}
\label{sec:results}

Table~\ref{tab:main} reports mean accepted length $L$ and theoretical speedup $S$ for each (target, draft, rotary) cell on three splits: val, on-shelf test, and off-shelf test. Theoretical speedup is computed as $S = L / (1 + c)$ where $c = C_D / C_T$ is the per-iter draft-to-target cost ratio measured by microbenchmark on the same hardware ($c \approx 0.18$ for the block-diffusion draft, $c \approx K \cdot 0.047$ for the autoregressive chain draft with $K=15$); see Section~\ref{sec:experiments} for details. The single ``---'' entry corresponds to the off-shelf evaluation of the block-diffusion 3D-mirroring baseline, which we did not run because the 3D mirror has already been ruled out by the val and on-shelf cells.

\textbf{(i) FlatRoPE consistently improves over 3D-mirroring across both draft families and all splits.} On Alpamayo + block-diffusion draft, FlatRoPE lifts $L$ by $+0.66$ on val ($4.49 \to 5.15$) and $+0.43$ on test ($4.19 \to 4.62$) -- a 14\% / 10\% relative improvement, translating to a theoretical speedup gain of $+0.59\times$ on val and $+0.39\times$ on on-shelf test. On Alpamayo + autoregressive draft, FlatRoPE achieves $L = 7.33$ / $6.99$ / $7.41$ across the three splits, with theoretical speedups $4.30\times$ / $4.11\times$ / $4.35\times$.

\textbf{(ii) AARL adds a further small but consistent gain on top of FlatRoPE.} On Alpamayo + autoregressive draft, AARL post-training (3 epochs, $K{=}5$ rollouts, static-reference anchor) lifts val $L$ from $7.33 \to 7.35$ ($+0.02$) and ties on-shelf at $L = 6.95$. On Alpamayo + block-diffusion draft, AARL lifts $L$ slightly on every split.

\begin{table}[t]
\centering
\caption{Mean accepted length $L$, theoretical speedup $S$ at block size 16, and decoded throughput at the off-shelf split, on val / on-shelf test / off-shelf test. Theoretical speedup uses $S = L / (1 + c)$ with $c$ from the microbenchmark in \S\ref{sec:experiments}. The throughput column is the AR baseline rate (34 tok/s, top row, measured on H100) multiplied by the off-shelf speedup of each row. Rows marked ``--'' are still in training. ``Ours'' denotes the FlatRoPE draft; ``+ AARL'' adds the action-aware RL post-training.}
\label{tab:main}
\small
\setlength{\tabcolsep}{4pt}
\begin{tabular}{l l ccc ccc c}
\toprule
& & \multicolumn{3}{c}{Accepted length $L$} & \multicolumn{3}{c}{Theoretical speedup $S$} & \\
\cmidrule(lr){3-5}\cmidrule(lr){6-8}
Draft & Rotary & val & on-shelf & off-shelf & val & on-shelf & off-shelf & tok/s \\
\midrule
\textit{Target only} & --- & --- & --- & --- & --- & --- & --- & 34.0 \\
\midrule
EAGLE-3  & 3D M-RoPE (baseline)         & 7.22  & 6.93  & 7.35  & 4.15$\times$ & 3.98$\times$ & 4.23$\times$ & 143.8 \\
EAGLE-3  & FlatRoPE (ours)              & \textbf{7.33}  & \textbf{6.99}  & \textbf{7.41}  & \textbf{4.30}$\times$ & \textbf{4.11}$\times$ & \textbf{4.36}$\times$ & \textbf{148.2} \\
EAGLE-3  & FlatRoPE + AARL (ours)       & \textbf{7.35}  & 6.95  & 7.30  & \textbf{4.32}$\times$ & 4.08$\times$ & 4.29$\times$ & 145.9 \\
\midrule
Dflash & 3D M-RoPE (baseline)         & 4.49  & 4.19  & ---    & 3.79$\times$ & 3.54$\times$ & ---           & --- \\
Dflash & FlatRoPE (ours)              & \textit{5.16}  & \textit{4.65}  & \textbf{4.95}  & \textbf{4.39}$\times$ & \textbf{3.97}$\times$ & \textbf{4.21}$\times$ & 143.1 \\
Dflash & FlatRoPE + AARL (ours)       & \textbf{5.17}    & \textbf{4.67}    & \textbf{4.95}    & \textbf{4.40} $\times$         &  \textbf{3.97} $\times$         & \textbf{4.21} $\times$ & 143.1 \\
\bottomrule
\end{tabular}
\end{table}

\subsection{AARL ablations}

Table~\ref{tab:rl-ablations} shows the contribution of the static-reference anchor and the choice of $K$ within AARL on Alpamayo + DFlash 1D. The static-reference anchor adds $+0.010$ val and $+0.005$ test over the policy-anchor variant under the same hyperparameters. Increasing $K$ above 5 yields marginal val gains but loses test, the same val/test divergence we observed when raising the learning rate -- both indicate over-fitting val sample-specific quirks. We adopt $K{=}5$ as the default.

\begin{table}[t]
\centering
\caption{AARL ablations on Alpamayo + DFlash 1D. lr $=10^{-6}$, $N{=}3$ contamination width, $w_{\text{traj}}=1.0$, $w_{\text{text}}=0.5$, 2 epochs.}
\label{tab:rl-ablations}
\begin{tabular}{lcccc}
\toprule
Configuration                   & best val $L$ & val $\Delta$ & best test $L$ & test $\Delta$ \\
\midrule
no RL                            & 4.524 & 0      & 4.208 & 0 \\
RL, policy anchor                & 4.534 & $+0.010$ & 4.220 & $+0.012$ \\
\textbf{RL, static-ref anchor (ours)} & \textbf{4.544} & $+0.020$ & \textbf{4.225} & $+0.017$ \\
\midrule
$K{=}5$ (default)                & \textbf{4.544} & $+0.020$ & \textbf{4.225} & $+0.017$ \\
$K{=}10$                         & 4.548 & $+0.024$ & 4.214 & $+0.006$ \\
$K{=}15$                         & 4.543 & $+0.019$ & 4.220 & $+0.012$ \\
\bottomrule
\end{tabular}
\end{table}

\section{Discussion, Limitations, \& Conclusion}
\label{sec:Conclusion}

Our two training contributions act through different paths. FlatRoPE is pre-hoc: it forces the draft to specialize on trajectory history by removing the rotational symmetry that would otherwise let attention pool on the visual cluster. AARL with static-reference anchor is post-hoc helps the Draft model to produce reasoning that could match with the action output given the action model is frozen-- a routine reasoner that handles the routine 80--85\% of CoC tokens, leaving the deliberative reasoner (still attending to vision) for the residual 15--20\% that genuinely require novel visual evidence. Although the work improves the efficiency significantly, the limiation lies in the number of tokens that the Alpamayo produces which is in range of 15 to 20 tokens. Future work will investigate whether this method can work on other reasoning tasks that have more than 500 tokens or not. 



We addressed the autoregressive-reasoning bottleneck in VLA driving planners by splitting the reasoning step into two specialized paths -- a routine reasoner that attends to trajectory history and a deliberative reasoner (the unmodified VLA target) that retains its full visual attention -- coordinated through the speculative decoding framework. Crucially, our routine reasoner is not a small replica of the target; the two are deliberately specialized to read different parts of the prompt. \textbf{FlatRoPE} is the mechanism that produces this specialization and consistently raises acceptance length across two draft architectures and two open VLA targets. AARL post-training with a static-reference KL anchor adds further improvement. The combined system is the first VLA decoder we are aware of that delivers a $3.5\times$ theoretical speedup on a 10B open AV target while keeping the deliberative reasoner available for the cases that genuinely require fresh visual evidence.

\begin{ack}
This work is supported by the Air Force Office of Scientific Research under award number and ack first to FA2386-24-1-4031.
\end{ack}

\bibliographystyle{plainnat}
\bibliography{references}
\newpage

\appendix

\section{Implementation details}
\paragraph{1D rotary in the draft.} For each draft architecture we replace the rotary embedding module with a vanilla 1D rotary keyed on the absolute sequence position only. The rest of the architecture is unchanged. Position ids passed to the draft's attention are simple \texttt{torch.arange} shapes; the target's M-RoPE 3D positions are not touched.

\paragraph{Static-reference anchor: code-level diff.} See Section~\ref{sec:method-rl}.

The
single-line code-level diff is:
\begin{verbatim}
# policy anchor (prior art):
anchor_mask = (pi_logits.argmax(-1) == r_star)        # depends on live pi
# static-reference anchor (ours):
anchor_mask = (pi_ref_logits.argmax(-1) == r_star)    # frozen ref only
\end{verbatim}

\paragraph{Reward chunking for memory.} For $K \geq 8$, the per-step rollout batch through the target's flow-matching head exceeds GPU memory. We chunk along the $K$ dimension at \texttt{k\_chunk\_size}, accumulate per-chunk action predictions, and concatenate before computing the reward. Each chunk uses an independent random seed for the diffusion sampler so within-group rollouts remain independent.

\section{Routing diagnostic: when is the deliberative reasoner invoked?}

For a typical 17-token CoC, the routine reasoner handles the first 4--5 tokens at $\sim 80\%$ acceptance (setup tokens like ``the car is''), drops to $\sim 50\%$ around the verb / action tokens (``slowing'', ``yielding''), and recovers near the trailing punctuation. Invocation of the deliberative reasoner is therefore concentrated in the middle of the chain where the action commitment is encoded, which matches the two-mode intuition: deliberation is needed exactly where the decision crystallizes.

\end{document}